\documentclass[letterpaper, 10 pt, conference]{ieeeconf}  

\IEEEoverridecommandlockouts                              

\usepackage{cite}
\usepackage{amsmath,amssymb,amsfonts}
\usepackage{mathtools}
\usepackage{algorithmic}
\usepackage{graphicx}
\usepackage{algorithm,algorithmic}
\usepackage{hyperref}
\hypersetup{hidelinks=true}
\usepackage{textcomp}
\newcommand{\vect}[1]{\mathbf{#1}}

\usepackage[super]{nth}

\usepackage{todonotes}
\usepackage{multirow}
\usepackage{siunitx}
\usepackage{tabularx}
\usepackage{comment}
\usepackage{float}
\usepackage{graphics} 
\usepackage{graphicx}
\usepackage{subcaption}
	
\usepackage[capitalise]{cleveref}
\usepackage{soul,xcolor}
\usepackage[T1]{fontenc}

\usepackage[firstpageonly=true]{draftwatermark}

\newcommand\norm[1]{\left\lVert#1\right\rVert}

\newcolumntype{Y}{>{\centering\arraybackslash}X}
\newcolumntype{Z}{>{\hsize=.05\hsize}X}

\newcommand*{\defeq}{\mathrel{\vcenter{\baselineskip0.5ex \lineskiplimit0pt
                     \hbox{\scriptsize.}\hbox{\scriptsize.}}}%
                     =}

\title{\LARGE \bf
Physics-Informed Neural Network \\for Multirotor Slung Load Systems Modeling
}

\author{Gil Serrano\textsuperscript{$\star$}, Marcelo Jacinto\textsuperscript{$\star$}, José Ribeiro-Gomes\textsuperscript{$\star$}, João Pinto\textsuperscript{$\star$}, \\
Bruno J. Guerreiro, Alexandre Bernardino, and
Rita Cunha
\thanks{\textsuperscript{$\star$}These authors contributed equally.}
\thanks{The work of G. Serrano, M. Jacinto, J. Ribeiro-Gomes, and J. Pinto was supported by the PhD Grants PRT/BD/154275/2022, 2022.09587.BD, 2022.12145.BD from MIT Portugal and Funda\c{c}{\~a}o para a Ci{\^e}ncia e a Tecnologia
(FCT), Portugal. This work was also supported by FCT, Portugal through LARSyS [DOI: 10.54499/LA/P/0083/2020, 10.54499/UIDP/50009/2020, and 10.54499/UIDB/50009/2020], project CAPTURE [DOI: 10.54499/PTDC/EEI-AUT/1732/2020], and project HAVATAR [DOI: 10.54499/PTDC/EEI-ROB/1155/2020].}
\thanks{The authors are with the Institute for Systems and Robotics (ISR), Laboratory of Robotics and Engineering Systems (LARSyS), Instituto Superior Técnico, University of Lisbon, Portugal. B. J. Guerreiro is also with CTS/Uninova, NOVA School of Science and Technology (FCT-NOVA), Caparica, Portugal. E-mails: {\tt \small \{gil.serrano, josepgomes\}@tecnico.ulisboa.pt}, {\tt \small \{mjacinto, jpinto, alex, rita\}@isr.tecnico.ulisboa.pt}, {\tt \small bj.guerreiro@fct.unl.pt}}}

\begin{document}

\SetWatermarkText{This paper has been accepted for publication in the\\ 2024 IEEE International Conference on Robotics and Automation (ICRA)}
\SetWatermarkColor[gray]{0.3}
\SetWatermarkFontSize{0.5cm}
\SetWatermarkAngle{0}
\SetWatermarkVerCenter{1.0cm}

\maketitle
\thispagestyle{empty}
\pagestyle{empty}

\begin{abstract}
Recent advances in aerial robotics have enabled the use of multirotor vehicles for autonomous payload transportation. Resorting only to classical methods to reliably model a quadrotor carrying a cable-slung load poses significant challenges. On the other hand, purely data-driven learning methods do not comply by design with the problem's physical constraints, especially in states that are not densely represented in training data. In this work, we explore the use of physics-informed neural networks to learn an end-to-end model of the multirotor-slung-load system and, at a given time, estimate a sequence of the future system states. An LSTM encoder-decoder with an attention mechanism is used to capture the dynamics of the system. To guarantee the cohesiveness between the multiple predicted states of the system, we propose the use of a physics-based term in the loss function, which includes a discretized physical model derived from first principles together with slack variables that allow for a small mismatch between expected and predicted values. To train the model, a dataset using a real-world quadrotor carrying a slung load was curated and is made available. Prediction results are presented and corroborate the feasibility of the approach. The proposed method outperforms both the first principles physical model and a comparable neural network model trained without the physics regularization proposed.

\end{abstract}

\vspace{0.2cm}
\urlstyle{same}
\noindent\textbf{Code + Dataset:} \href{https://github.com/GilSerrano/pinn-air}{https://github.com/GilSerrano/pinn-air} \\
\noindent\textbf{Video:} \href{https://youtu.be/4N\_nnaGsUOg}{https://youtu.be/4N\detokenize{_}nnaGsUOg}


\section{Introduction}
\label{sec:introduction}


System identification is a fundamental task in model-based control of dynamical systems, as it involves the estimation of the underlying dynamics from input-output data -- due to either the model being hard to derive or its parameters being unknown \textit{a priori}. 
Characterizing a physical system through observed data is critical to understanding its dynamics and allowing the design of controllers that enable the system to follow reference signals with a certain desired behavior.

Classical approaches to system identification mainly rely on the statistical analysis of the input-output relation~\cite{ljung1999system, soderstrom1989system, van2012subspace} to extract crucial information from collected data. Time and frequency-domain analyses are some of the more common approaches. Using a pseudo-random binary signal (PBRS) that excites most frequencies, one can, with relative ease, design linear auto-regressive models that capture the most predominant modes of the system. In contrast, neural networks with nonlinear activations are able to learn nonlinear approximations of complex system dynamics.

Recent and promising approaches have attempted to use neural networks to model unknown system dynamics. Physics-Informed Neural Networks (PINNs~\cite{PINNs}) are neural networks that are trained to solve supervised learning tasks while taking into account the physical constraints of the world. By exploiting the knowledge of the physical laws that govern the system under analysis in the learning process, PINNs overcome the low data availability that may occur when dealing with real systems. This prior knowledge acts as a regularization term that limits the space of admissible solutions. Other approaches resort to a combination of Recursive Neural Networks (RNNs) with variational auto-encoders\cite{Gedon_DeepStateSpace} or hybrid physical models \cite{RNN_HYBRID,MixNet} to capture the recursive nature of dynamical systems.



\begin{figure*}
    \centering
    \includegraphics[width=\textwidth]{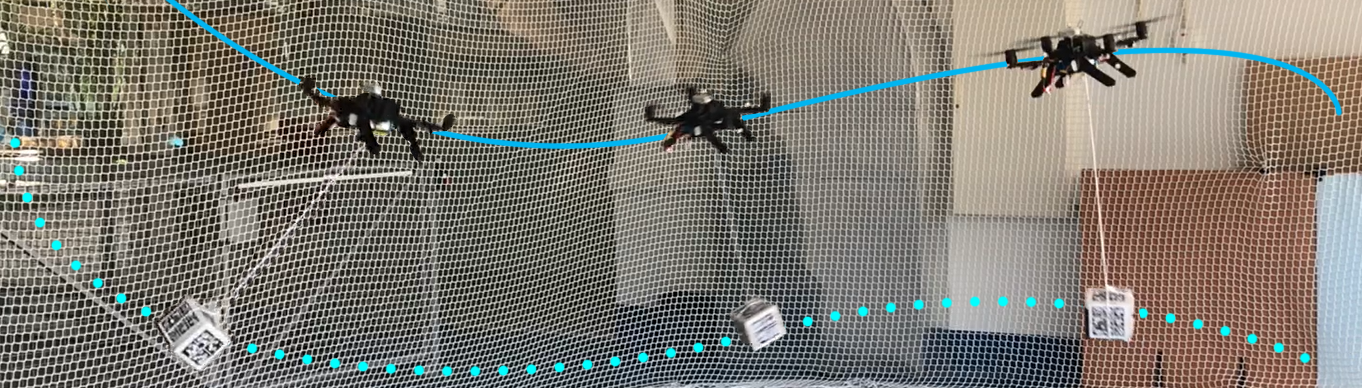}
    \caption{Intel Aero quadrotor carrying a cube-shaped slung load attached by a string in a dedicated motion capture arena.}
    \label{fig:project_description}
\end{figure*}

The combination of data-driven methods with physically plausible models is specially important in systems with large degrees of uncertainty. For instance, the interaction between a multirotor aircraft and a slung load (Fig.~\ref{fig:project_description}) makes system identification challenging, due to phenomena such as downwash on the load. While this system is relatively simple, it has interesting use cases such as aerial transportation \cite{9195317}. Classical approaches for system identification may not be well-suited for this problem, as they struggle to accurately capture the nonlinear dynamics of the system over long time periods, since the maneuvers that can be performed to excite the modes of the coupled system (drone and slung load) are limited. Deep learning approaches, on the other hand, have shown promise in deducing complex dynamics and how they evolve over time \cite{sindy_nn, PINNs,PINN_control_antonelo,pinns_mpc_robotic_arm, Gedon_DeepStateSpace}. The development of controllers that leverage learned dynamics could be a significant step toward the successful operation of multirotor systems carrying slung loads.

\subsubsection*{Contributions}
This work proposes the use of a Long-Short Term Memory (LSTM) encoder-decoder network with an attention mechanism to model end-to-end the system dynamics and perform multi-step predictions of a quadrotor carrying a slung load. The main contributions are:
\begin{itemize}
    \item an encoder-decoder architecture to solve the prediction problem making use of slack variables and a physics-based loss term -- acting as a regularization term that limits the space of search and designed by taking into account a discrete approximation of the system derived from first principles;
    \item a novel real-world dataset collected in a motion capture (MOCAP) arena that was obtained by conducting 48 independent flight maneuvers. 
\end{itemize}

The paper is organized as follows: \cref{sec:related_work} summarizes the research conducted on the intersection of deep learning and system identification for the control of dynamical systems; \cref{sec:proposed_method} presents the problem formulation, with a focus on the multirotor dynamics, and the proposed network architecture and loss function; \cref{sec:num_results} details the data acquisition and pre-processing required to then train the network, the evaluation metrics employed, and presents the results obtained. Concluding remarks are offered in \cref{sec:conclusion} along with a future line of research. 


\section{Related Work}
\label{sec:related_work}

Recent approaches have attempted to use neural networks to model unknown system dynamics.  One such approach is the Sparse Identification of Nonlinear Dynamics (SINDy)~\cite{sindy} framework, which proposes expressing the dynamics, given by ordinary differential equations (ODEs), as a linear combination of nonlinear basis functions. To determine the contribution of each basis function, a convex sparse linear regression problem is formulated. An extension of SINDy proposes combining this idea with the use of a neural network with an autoencoder structure to simultaneously discover a low-dimensional latent space of coordinates and functions that model the dynamics of the system~\cite{sindy_nn}. 

Another concept in system identification is Physics-Informed Neural Networks (PINNs) \cite{PINNs}. These networks are capable of encoding the fundamental dynamics that govern a dataset, 
given measurements of the evolution of the system described by a nonlinear partial differential equation (PDE) $\partial_t g + \mathcal{N}[g; \lambda] = 0$, where $g(x,t)$ denotes the latent solution, dependent on variables $x$ and $t$, and $\mathcal{N}[g;\lambda]$ is a nonlinear differential operator parameterized by a vector $\lambda$. By minimizing the sum of the MSE (Mean Squared Error) of the boundary conditions and the MSE of the fit of a set of points in $(x, t)$ with regards to the differential equation, PINNs are able to estimate the parameters $\lambda$ that best describe the observed data. Standard PINN models are useful in a variety of problems, however, they hinge on the knowledge of $\mathcal{N}$ and are not suited for control tasks, as they are not designed to handle variable control inputs.

Some work has been done to make PINNs useful in control applications. In \cite{PINN_control_antonelo}, Physics-Informed Neural Nets for Control (PINCs) were introduced, which extend PINNs by adding control actions as inputs. PINCs are able to identify a dynamical system from data and its governing ODEs, and can also simulate differentiable equations faster than conventional numerical methods. This approach is able to propagate the equations for variable horizons without serious degradation in the predictions, measured with absolute error, which is critical for model predictive control applications.

The research presented in \cite{pinns_mpc_robotic_arm} also extended PINNs by adding control actions as network inputs. With these modifications, the authors were able to combine PINNs with Nonlinear Model Predictive Control (NMPC). At each instant, the neural network receives the current time step, state, and input and predicts the next state, thus effectively replacing an explicit model in the NMPC algorithm. Simulations demonstrate that this approach can solve a tracking problem for a robotic manipulator, surpassing classical time-integration methods in terms of computational efficiency. The performance is evaluated with the Root Mean Squared Error (RMSE) along the reference trajectory.

In \cite{Gedon_DeepStateSpace}, Deep State Space (DSS) models are presented as a system identification approach that can describe a wide range of dynamics due to the flexibility of deep neural networks. The focus of that work is on sequential data and the proposed strategies are based on RNNs, as they are useful in modeling sequences of variable length, and variational autoencoders. 

The approach presented in this work is inspired by the original work on PINNs \cite{PINNs} and DSS \cite{Gedon_DeepStateSpace} with the key difference being the combination of RNNs with an encoder-decoder structure to model a discretized dynamical system with variable control inputs, and the use of a physical model derived from first principles applied to the network output during training to restrict the space of admissible predictions, combined with auxiliary slack variables that enhance the ability to account for unmodeled system dynamics.


\section{Proposed Method}
\label{sec:proposed_method}

The dynamics of a quadrotor carrying a slung load are governed by a differential equation that can be derived as in Yu et al. \cite{slung_load_yu_gan}, where it is assumed that the load is rigidly connected to the center of mass of the quadrotor by a fixed-length cable that is taut during flight. The dynamical model adopted does not account for aerodynamic drag, the downwash on the load, structure bending modes, nor external disturbances, which can prove hard to quantify accurately in practice. In this section, we present the physical model in its discrete form. The network that is described in \cref{sec:net_architecture} uses this knowledge during training to learn the dynamics of the full system composed of the quadrotor and a slung load.

\subsection{System model}
\label{sec:system_model}
Let $\mathcal{W}$ denote the world reference frame, which corresponds to the canonical basis of $\mathbb{R}^3$, $\mathcal{W} = \{\vect{e}_i\}_{i = 1,2,3}$, and $\mathcal{B}$, the body frame attached to the vehicle with origin at the center of mass of the quadrotor. The position $\vect{p}^{\mathrm{L}}$ and velocity $\vect{v}^{\mathrm{L}}$ of the payload in $\mathcal{W}$, considering a sampling time $h \in \mathbb{R}^+$, at time instant $k+1$ are approximated by
\begin{equation}
\begin{cases}
	\vect{p}^{\mathrm{L}}_{k+1} &= \vect{p}^{\mathrm{L}}_{k} + h \vect{v}^{\mathrm{L}}_k + \frac{1}{2} h^2 \vect{u}^{\star}_k \\
	\vect{v}^{\mathrm{L}}_{k+1} &= \vect{v}^{\mathrm{L}}_{k} + h \vect{u}^{\star}_k
\end{cases}~,
\end{equation}
with the virtual input $\vect{u}^{\star}_k$ given by
\begin{equation}
	\vect{u}^{\star}_k = -\frac{1}{m_\mathrm{L}}T^{\mathrm{L}}_k \boldsymbol{\xi}_k + g\vect{e}_3 ~,
\end{equation}
where $m_\mathrm{L} \in \mathbb{R}^+$ denotes the mass of the load, $T^\mathrm{L}_k \in \mathbb{R}$ is the tension in the cable, $\boldsymbol{\xi}_k \in \mathbb{S}^2$ is the load cable direction, and $g = \SI{9.8}{\meter\per\second\squared}$ denotes Earth's gravity. The evolution of the position $\vect{p}$ and velocity $\vect{v}$ in $\mathcal{W}$ of the multirotor are provided by 
\begin{equation}
\begin{cases}
	\vect{p}_{k+1} &= \vect{p}_k + h \vect{v}_k + \frac{1}{2} h^2 (\overline{\vect{u}}_k + \vect{u}^{\star}_k) \\
	\vect{v}_{k+1} &= \vect{v}_k + h (\overline{\vect{u}}_k + \vect{u}^{\star}_k)
\end{cases}~,
\end{equation}
where the virtual input $\overline{\vect{u}}_k$ is given by
\begin{equation}
	\overline{\vect{u}}_k = \frac{T^\mathrm{Q}_k}{m_\mathrm{Q}} \vect{R}_k \vect{e}_3 - \frac{m_\mathrm{Q} + m_\mathrm{L}}{m_\mathrm{Q}}g\vect{e}_3 ~,
\end{equation}
where $m_\mathrm{Q} \in \mathbb{R}^+$ denotes the mass of the vehicle, $T^\mathrm{Q}_k \in \mathbb{R}^+$ corresponds to the total thrust produced by the rotors, and $\vect{R}_k \in \mathrm{SO(3)}$ corresponds to the vehicle orientation. The orientation kinematics are propagated as follows,
\begin{equation}
    \vect{R}_{k+1} = \vect{R}_k \exp{ \{ h \hspace{.5ex} \vect{S}(\boldsymbol{\omega}_k) \} } ~,
\end{equation}
with $\boldsymbol{\omega}_k\in\mathbb{R}^3$ representing the quadrotor angular velocity expressed in the body frame $\mathcal{B}$, and $\vect{S}(\cdot)$ is an isomorphism between $\mathbb{R}^3$ and the space of $3 \times 3$ skew-symmetric matrices defined by the condition $\vect{S}(\vect{a})\vect{b} = \vect{a} \times \vect{b}$, $\forall\vect{a},\vect{b} \in \mathbb{R}^3$. The load cable direction vector can be obtained by
\begin{equation}
	\boldsymbol{\xi}_k = \frac{\vect{p}^{\mathrm{L}}_{k} - \vect{p}_{k}}{\ell},
\end{equation}
where $\ell$ denotes the cable length. The tension on the cable can be computed according to
\begin{equation}
	T^{\mathrm{L}}_k = -\frac{m_\mathrm{L}}{m_\mathrm{L} + m_\mathrm{Q}} \boldsymbol{\xi}_k^{\top}\vect{R}_k T^{\mathrm{Q}}_k \vect{e}_3 + \frac{m_\mathrm{L} m_\mathrm{Q}}{m_\mathrm{L}+m_\mathrm{Q}} \ell \norm{\boldsymbol{\Omega}^{\mathrm{L}}_k}^2 ~,
\end{equation}
where $\boldsymbol{\Omega}^{\mathrm{L}}_k \in \mathbb{R}^3$ is the angular velocity of the load and its temporal evolution is determined by
\begin{equation}
	\boldsymbol{\Omega}^{\mathrm{L}}_{k+1} = (1 - d) \boldsymbol{\Omega}^{\mathrm{L}}_{k} - \frac{h}{m_\mathrm{Q} \ell} \vect{S}(\boldsymbol{\xi}_k) \vect{R}_k T^{\mathrm{Q}}_k \vect{e}_3~,
\end{equation}
with $d \in [0, 1]$, a damping term. The observed variables of the system state are stored on the tuple $\vect{x}_k = \left\{ \vect{p}_k, ~\vect{v}_k, ~\vect{q}_k, ~\vect{p}_k^\mathrm{L} \right\}$, corresponding to the position, linear velocity, and orientation -- expressed using unit quaternions $\vect{q}_k \in \mathbb{S}^3$ -- of the vehicle, and the position of the load $\vect{p}_k^{\mathrm{L}}$ expressed in $\mathcal{W}$. The term $\boldsymbol{\Omega}^{\mathrm{L}}_{k}$ is not included in $\vect{x}_k$, as it is not measured directly. The orientation is represented as a unit quaternion because this parameterization has been shown to converge faster and produce better results than rotation matrices in the context of neural networks~\cite{zhou2019continuity}. The control input of the complete system is given by $\vect{u}_k = \{ T^{\mathrm{Q}}_k, ~\boldsymbol{\omega}_k \}$ and is comprised of the total thrust produced by the quadrotor and its body angular rates. The evolution of $\vect{x}_k$ is, therefore, described by $\vect{{x}}_{k+1} = f(\vect{{x}}_{k}, \vect{u}_{k}, \boldsymbol{\Omega}^{\mathrm{L}}_{k})$. 




\subsection{Network architecture}
\label{sec:net_architecture}

\begin{figure*}
    \centering
    \includegraphics[width=\textwidth]{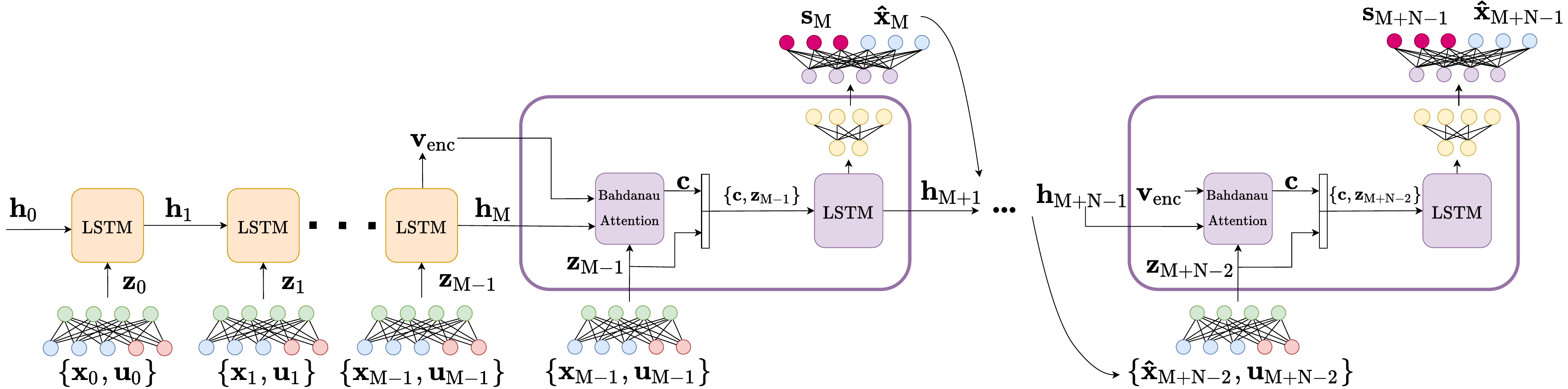}
    \caption{Proposed encoder-decoder architecture with an attention mechanism. The network encoder receives a temporal history of $\mathrm{M}$ system states and control inputs. The decoder receives the hidden state and output of the encoder, the last measured state of the system $\vect{x}_{\mathrm{M-1}}$, and the previous input of the system $\vect{u}_{\mathrm{M-1}}$, which are combined in an attention module to produce a prediction of the next system state $\vect{\hat{x}}_{\mathrm{M}}$ and the corresponding slack variables $\vect{s}_{\mathrm{M}}$. Each prediction of the decoder module is concatenated with an input control signal to generate multi-step predictions of the system state in the following iterations.}
    \label{fig:encoder_decoder}
    \vspace{-0.4cm}
\end{figure*}

Given a time series of observations of state variables and control inputs $\{\vect{x}_k, \vect{u}_k \}$ up to time step $\mathrm{M}-1$, the network aims to recursively predict the evolution of the state of the quadrotor-slung-load system for the next $\mathrm{N}$ time steps, given only the control input that is applied to the vehicle at those instants. The architecture of the proposed network is shown in \cref{fig:encoder_decoder}. We start by providing $\{\vect{x}_k, \vect{u}_k \}$ to an encoder -- these vectors are passed through fully connected layers with GELU activations  to discover a latent space representation of the input data $\{ \vect{z}_k \}, ~k = 0, 1, \hdots, \mathrm{M}-1$. This sequence of new feature vectors is fed to a multilayer LSTM to produce a hidden state $\vect{h}_\mathrm{M}$ and the encoder output $\vect{v}_{\mathrm{enc}}$. 

To obtain the estimates of the system state $\vect{\hat{x}}_k$ in the subsequent time steps, we use a decoder composed of a multilayer LSTM network with a Bahdanau attention mechanism~\cite{Bahdanau2014}. At $k=\mathrm{M}$, the decoder receives the encoder output $\vect{v}_{\mathrm{enc}}$, the previous hidden state $\vect{h}_{\mathrm{M}}$, and the last known system state and the actual control input $\{\vect{x}_{\mathrm{M}-1},\vect{u}_{\mathrm{M}-1}\}$. The system state and the control input are fed to fully connected layers and the output $\vect{z}_{\mathrm{M}-1}$ is passed to the attention layer, along with the encoder output and previous hidden state. Then, the context $\vect{c}$ that is calculated by the attention mechanism is concatenated with $\vect{z}_{\mathrm{M}-1}$ and fed to the LSTM. The output of the LSTM is then passed through fully connected layers that convert it to the original state space, thus obtaining the estimate $\vect{\hat{x}}_\mathrm{M}$, which is concatenated with the next actual input to make a recursive prediction. Additionally, a set of slack variables $\vect{s}_\mathrm{M}$ that is used by the loss function to impose soft-physical constraints on the system predictions during training is generated. In the following time steps, $k = \mathrm{M}+1, ..., \mathrm{M+N}-1$, the previous system state estimate $\vect{\hat{x}}_{k-1}$ is fed to the fully connected layers, instead of the actual system state, along with the control input. 

To train the model, a loss function given by the sum of four separate terms is considered
\begin{equation}
    \mathcal{L} = \mathcal{L}_{\mathrm{fit}} + \mathcal{L}_{\mathrm{physics}} + \mathcal{L}_{\mathrm{projection}} + \mathcal{L}_{\mathrm{slack}}.
\end{equation}
The first term of the loss is included to enforce the network to fit the training data,
\begin{equation} \label{eq:fit_loss}
    \resizebox{.425\textwidth}{!}{$
    \begin{split}
        \mathcal{L}_{\mathrm{fit}} = \sum_{n = \mathrm{M}}^{\mathrm{M}+\mathrm{N}-1} \lambda_1 \left\| \vect{\hat{p}}_n - \vect{p}_n \right\|_{\vect{M}_{n}}^2 + \lambda_2 \left\| \vect{\hat{v}}_n - \vect{v}_n \right\|_{\vect{M}_{n}}^2 \\
        +\, \lambda_3 \left\| \vect{e}(\vect{\hat{q}}_n, \vect{q}_n) \right\|_{\vect{M}_{n}}^2
        + \lambda_4 \left\| \vect{\hat{p}}_n^{\mathrm{L}} - \vect{p}_n^{\mathrm{L}} \right\|_{\vect{M}_{n}}^2,
    \end{split}$}
\end{equation}
\noindent 
{where $\forall \vect{y} \in \mathbb{R}^{m},\, \left\| \vect{y} \right\|_{\vect{M}_{n}}^2 \defeq \vect{y}^\top \vect{M}_{n} \vect{y}$, $\vect{M}_{n} \defeq \vect{I}\,e^{-\alpha\,(n-M)}$, with $\alpha>0$ and $\vect{I}$ being the identity matrix of appropriate dimension; the map ${\vect{e}:\mathbb{S}^3\times\mathbb{S}^3\to\mathbb{R}^4}$ is defined as \begin{equation}
    \vect{e}(\vect{q}_1, \vect{q}_2) = \vect{q}_1 \otimes \vect{q}_2^{-1} - \vect{q}_{\mathrm{id}},
    \label{eqn:quaternion_error}
\end{equation}
where $\vect{q}_{\mathrm{id}}$ denotes the identity quaternion, to measure the error between two quaternions representing the orientation of the vehicle. The second term,
\begin{equation} \label{eq:phy_loss}
    \resizebox{.425\textwidth}{!}{$
    \begin{split}
        \mathcal{L}_{\mathrm{physics}} =~ & \phi \left\|\vect{\hat{x}}_{\mathrm{M}}- f(\vect{x}_{\mathrm{M-1}},\vect{u}_{\mathrm{M-1}},\boldsymbol{\Omega}^{\mathrm{L}}_{\mathrm{M-1}}) + \vect{s}_{\mathrm{M}}  \right\|_{\vect{Q}_\mathrm{M}}^2 \\ 
        + &\sum_{n = \mathrm{M}}^{\mathrm{M}+\mathrm{N}-2} \phi \left\|\vect{\hat{x}}_{n+1}- f(\vect{\hat{x}}_{n},\vect{u}_{n},\boldsymbol{\Omega}^{\mathrm{L}}_{n})  + \vect{s}_{n+1}\right\|_{\vect{Q}_{n+1}}^2,
    \end{split}$}
\end{equation}
ensures the system dynamics restrict the admissible search space. The function $f$ represents the discretized drone dynamics introduced in \cref{sec:system_model} so that $\vect{\hat{x}}_{k+1}\approx\,f(\vect{\hat{x}}_{k}, \vect{u}_{k}, \boldsymbol{\Omega}^{\mathrm{L}}_{k})$. The term $\mathcal{L}_{\mathrm{physics}}$ acts, therefore, as a regularization mechanism that penalizes solutions that do not satisfy the system dynamics. However, since the physical model is not ideal, there are discrepancies between the discretized model predictions and observations gathered from the real system, causing the estimates to diverge increasingly as prediction horizons become longer. As such, the slack variables $\vect{s}_k$ are introduced as extra states predicted by the network\footnote{The slack variables are only used during the training phase, and their value is discarded at evaluation time.}, which allow for a small mismatch between the predictions of the physical model and the network. Moreover, in the penalty term, a sequence of diagonal matrices $\vect{Q}_n \defeq \vect{I}\,e^{-\beta\,(n-M)}$, $\beta>0$, that converge exponentially to the zero matrix is introduced to penalize the mismatch between estimates and model predictions in the first instants and gradually less further into the future.

To represent a rotation, a quaternion must have unit norm and the network should estimate it as such. The loss function incorporates, therefore, this constraint as follows,
\begin{equation} \label{eq:proj_loss}
        \mathcal{L}_{\mathrm{projection}} = \psi \sum_{n = \mathrm{M}}^{\mathrm{M}+\mathrm{N}-1}  \big\lvert 1 - \| \vect{\hat{q}}_n \| \big\rvert^2.
\end{equation}

Finally, it is desirable that the slack variables $\vect{s}_k$ are close to zero so that the network predictions follow the physical model as closely as possible. To explicitly account for such requirement, the last term of the loss function is given by
\begin{equation} \label{eq:slack_loss}
    \mathcal{L}_{\mathrm{slack}} = \rho \sum_{n = \mathrm{M}}^{\mathrm{M}+\mathrm{N}-1}  \left\| \vect{s}_n \right\|_{\vect{Q}_n}^2.
\end{equation}

\noindent
The hyperparameters $\lambda_1, ~\lambda_2, ~\lambda_3, ~\lambda_4$ in (\ref{eq:fit_loss}); $\phi$ in (\ref{eq:phy_loss});  $\psi$ in (\ref{eq:proj_loss}); and $\rho$ in (\ref{eq:slack_loss}) determine the relative weighting of each of the squared error terms in $\mathcal{L}_{\mathrm{fit}}$, $\mathcal{L}_{\mathrm{physics}}$, $\mathcal{L}_{\mathrm{projection}}$, and $\mathcal{L}_{\mathrm{slack}}$, respectively.

\section{Results}
\label{sec:num_results}
In this section, we describe the data acquisition and processing procedures. Experiments are conducted to assess the performance of the proposed model and training strategy.
\begin{figure*}
    \centering
    \includegraphics[width=0.79\textwidth]{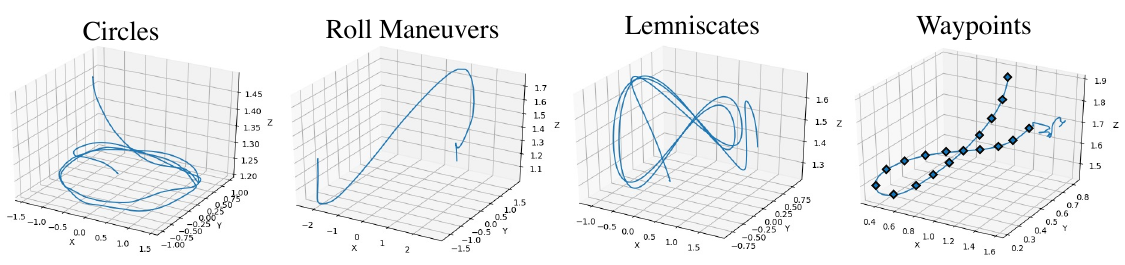}
    \caption{Subset of the trajectories executed by the quadrotor carrying a slung load during data acquisition.}
    \label{fig:flight_maneuvres}
    \vspace{-0.2cm}
\end{figure*}

\vspace{-1em}
\subsection{Data Acquisition and Model Training}
To acquire real-world data, a $\SI{100}{\gram}$ cube with $\SI{10}{\centi\meter}$ edges was attached to an Intel Aero RTF drone with a $\SI{60}{\centi\meter}$ string (\cref{fig:project_description}). The drone was commanded to perform various flight maneuvers to excite different modes of the system, as depicted in \cref{fig:flight_maneuvres}. The control signal $\vect{u}$ was determined by an offboard controller. The acquired data containing the state and input of the system was synchronized and sampled at  $h=\SI{30}{\milli\second}$. A total of 48 different experiments was conducted, with duration varying from 197 to 799 time steps. Each time series was divided into chunks of a fixed duration $\mathrm{M} = 50$ time steps, containing the state and control inputs to the system, and the model is trained to predict the state of the system for the next $\mathrm{N} = 25$ time steps. Additional evaluation tests were also conducted considering $\mathrm{N} = 50$.

A sliding window approach with stride $1$ was used to generate sequences from the flight data recorded. A split of 58-17-25 was adopted, corresponding to 9,959 sequences for training, 2,837 for validation, and 4,252 for testing. Each model is trained using Adam \cite{Kingma2014} with a learning rate of $1\times 10^{-4}$ and an $\ell_2$ weight decay of 0.05. The training procedure is performed for 3,000 epochs, with early-stopping and batches composed of 256 sequences on an NVIDIA RTX 3090 GPU, taking $\SI{6}{\second}$ per epoch, and a total of $\SI{5}{\hour}$ per model trained. The hyperparameters $\lambda_1=\lambda_2=\lambda_3=\lambda_4=20$, $\phi=5$, $\psi=10$, $\rho=0.1$, $\alpha=0.1$, $\beta=0.6$ were considered, and the trained model has 238,906 parameters. 

\subsection{Performance Metrics}
To assess the performance of the proposed model
and training strategy, we compare our model with a physics-based open-loop state predictor, derived from first principles, i.e. the model in Section \ref{sec:system_model}. Additionally, a performance comparison with a baseline LSTM encoder-decoder network trained on the same training data, but without slack variables and without the $\mathcal{L}_{\mathrm{physics}}$ and $\mathcal{L}_{\mathrm{slack}}$ loss terms is conducted -- enabling the evaluation of the impact of the physics constraints on the performance of the model. The performance of each model is analyzed by computing the evolution of the Mean Average Error (MAE) over all the test set sequences at each timestep $k$ according to
\begin{equation}
    \text{MAE}_k = \frac{1}{\mathrm{\beta}}\sum_{j=1}^{\mathrm{\beta}}\norm{\vect{\zeta}_k^j - \vect{\hat{\zeta}}_k^j},
\end{equation}
where $\vect{\zeta}_k^j$ and $\vect{\hat{\zeta}}_k^j$ are a subset of the groundtruth and prediction of the system state for the test sample $j$, respectively, and $\mathrm{\beta}$ is the number of test sequences. Note that the quaternion error is computed according to the formula presented in \eqref{eqn:quaternion_error}. The RMSE is additionally computed for each model over the entire test set.

\subsection{Prediction Performance}
In \cref{fig:std_avg_proposed_model} the evolution of the MAE over time for the proposed neural network with the discretized physical model as well as the Q1 and Q3 quartiles of the quadrotor position, velocity, orientation, and payload position are shown. From these results, it is possible to conclude that in the initial prediction steps, the error is smaller for the physics-based predictor, but as the prediction time increases, the neural network outperforms the classical model and its standard deviation increases at a slower rate. The dashed vertical line denotes the \nth{25} step of the prediction horizon adopted during training. The plots also demonstrate that even though the model is only trained to make predictions for $\SI{750}{\ms}$, it can be used over longer prediction horizons at evaluation time. 
\begin{figure}
    \centering
    \includegraphics[width=1.0\linewidth]{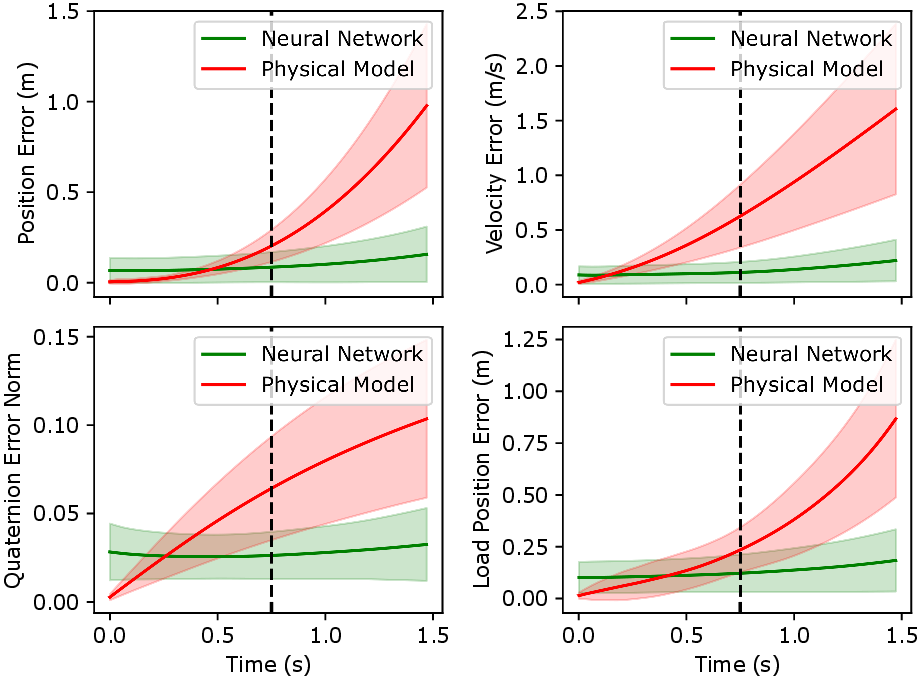}
    \caption{Comparison between the evolution of the MAE and standard deviation of the prediction error of the proposed model and the physics-based state predictor, evaluated on the test set. The vertical dashed line denotes the prediction horizon of $\mathrm{N}=25$ time steps ($\SI{750}{\ms}$) used during training.}
    \label{fig:std_avg_proposed_model}
    \vspace{-0.6cm}
\end{figure}
In \cref{fig:prediction_results} we showcase the results of the recursive prediction of the state of the system for a single sequence of the test set. The first $\mathrm{M}=50$ time steps ($\SI{1.5}{\second}$) are used as the initial input to the neural network, and the output of the proposed model is depicted in dashed lines for the following timesteps over the groundtruth. The time series predicted by the discretized dynamical model are depicted in dotted lines. In this particular example, it is possible to conclude from visual inspection that the learned model can accurately follow the evolution of each state during the first $\mathrm{N}=25$ prediction steps, while the predictions generated by the dynamical model start to deviate from the groundtruth, particularly the velocity predictions. This is due to the fact that the discretized dynamical model does not account for disturbances such as aerodynamic drag and the downwash on the slung load, in contrast to the network which is able to learn the unmodeled dynamics.
\begin{figure}
    \centering
    \includegraphics[width=1.0\linewidth]{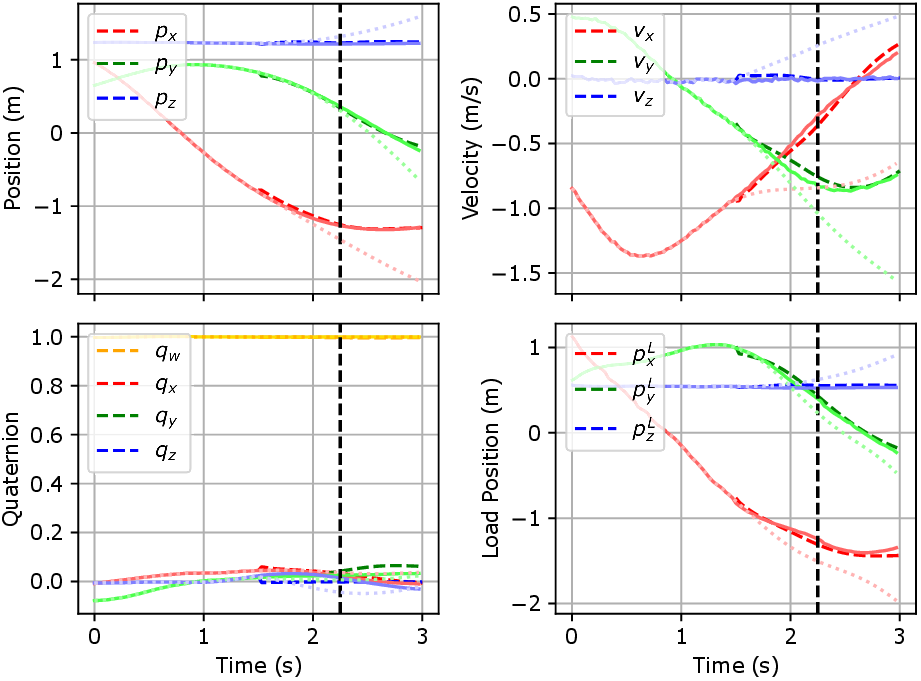}
    \caption{Prediction of the vehicle position, linear velocity, orientation, and the position of the slung load as a function of time for a single sequence of the test set. The time series predicted by the network are depicted in \textbf{dashed lines} over the expected values in \textbf{solid lines}. The time series predicted by the discretized dynamics model are depicted in \textbf{dotted lines} for comparison. The vertical dashed lines denote the prediction horizon of $\SI{750}{\ms}$ used during training.}
    \label{fig:prediction_results}
    \vspace{-0.25cm}
\end{figure}
\begin{figure}
    \centering
    \includegraphics[width=1.0\linewidth]{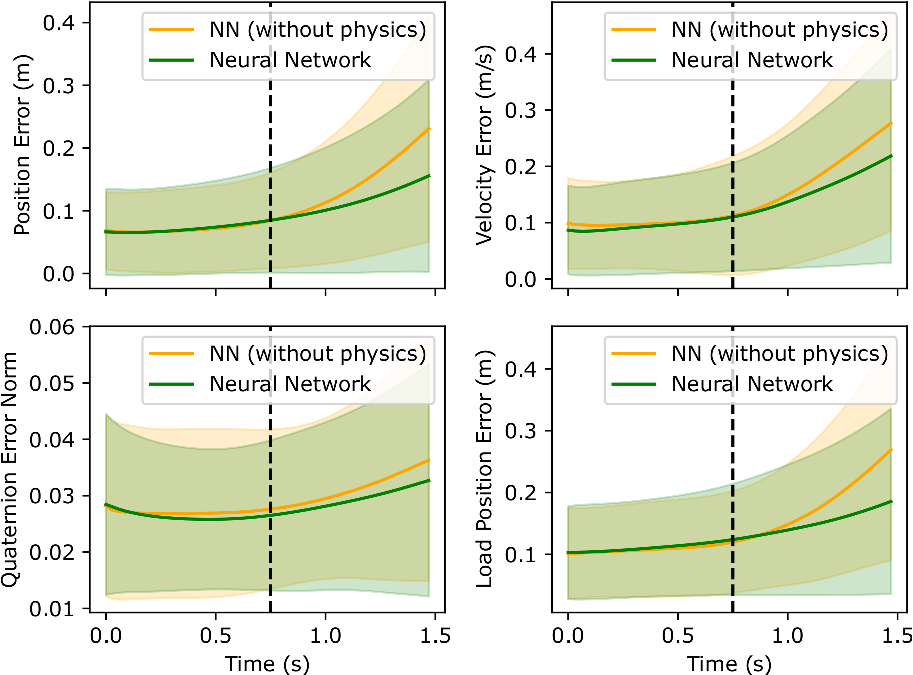}
    \caption{Evolution of the MAE and standard deviation of the prediction error of the proposed model with a baseline model trained without physics constraints. The vertical dashed line denotes the prediction horizon used during training.}
    \label{fig:std_avg_against_baseline}
    \vspace{-0.55cm}
\end{figure}

To evaluate the impact of imposing physical constraints on the loss function during training, \cref{fig:std_avg_against_baseline} compares the evolution of the MAE of the proposed model with an equivalent baseline model trained without the physical constraints. From these results, we confirm that enforcing physical constraints on the model reduces the mean prediction error over a longer prediction horizon, although the gains over the baseline neural network model are also shown to be smaller in comparison to the gains obtained over the physical model predictor. The RMSE results presented in \cref{tab:rmse_table} obtained over the entire test set, for the discretized physics model, and for the neural network trained with and without the physics loss term and slack variables further support the analysis. 
\begin{table}
    \centering
    \renewcommand{\arraystretch}{1.2}
    \caption{RMSE of the different models, over the entire test set for a prediction horizon of $\mathrm{N}=50$ time steps.}
    \resizebox{\columnwidth}{!}{%
    \begin{tabular}{ c|c|c|c|c|c } 
    \hline
    \multirow{3}{*}{Model} & \multicolumn{5}{c}{RMSE} \\ \cline{2-6}
     & \multirow{2}{*}{Position (m)} & \multirow{2}{*}{Velocity (m/s)} & \multirow{2}{*}{Quaternion} & Payload & \multirow{2}{*}{Combined} \\
     &  &  &  &  Position (m) &  \\
    \hline
    Physics & 0.458 & 0.924 & 0.073 & 0.424 & 1.879 \\
    \hline
    Proposed Model (w/o physics) & 0.153 & 0.193 & 0.033 & 0.182 & 0.561 \\
    \hline
    Proposed Model (w/o slack vars) & 0.153 & 0.200 & 0.034 & 0.184 & 0.571 \\
    \hline
    Proposed Model & \textbf{0.136} & \textbf{0.175} & \textbf{0.032} & \textbf{0.166} & \textbf{0.508} \\
    \hline
    \end{tabular}}
    \label{tab:rmse_table}
\end{table}
\noindent The inference times for different horizons $\mathrm{M}$ and $\mathrm{N}$ on an NVIDIA RTX 3090 GPU and on an NVIDIA Jetson Orin Nano are presented in \cref{tab:inference_time} and support the feasibility of the proposed model. 
\begin{table}
    \centering
    \renewcommand{\arraystretch}{1.2}
    \small
    \caption{Inference time.}
    \resizebox{0.8\columnwidth}{!}{%
    \begin{tabular}{ c|c|c|c|c|c } 
    \hline
    \multicolumn{2}{c|}{Horizon} & \multicolumn{2}{c|}{RTX 3090} & \multicolumn{2}{c}{Jetson Orin Nano} \\
    \hline
    $M$ & $N$ & mean (ms) & std dev (ms) & mean (ms) & std dev (ms)\\ 
    \hline
    50 & 25 & 10.76 & 0.53 & 59.44 & 0.26 \\
    50 & 50 & 21.24 & 0.99 & 115.26 & 0.59\\
    \hline
    \end{tabular}}
    \label{tab:inference_time}
    \vspace{-0.4cm}
\end{table}

\vspace{-0.15cm}
\section{Conclusion}
\label{sec:conclusion}

We have proposed a network architecture to recursively predict the evolution of a quadrotor-slung-load system from observations of its state and control inputs. The network features an encoder-decoder structure with attention that provides estimates of the state of the quadrotor-slung-load system in subsequent time steps along with auxiliary slack variables that allow to better account for unmodeled system dynamics. Moreover, the modeled quadrotor-slung-load dynamics discounted over the prediction horizon are incorporated into the loss function to train the network so that more accurate predictions of the evolution of the coupled system are obtained.


The network was trained using data from various flight maneuvers performed in a controlled experimental setup. The results show the model is able to accurately predict how the dynamics of the system evolve in time, outperforming the predictions made by a model derived from first principles. The impact of the physics constraints and slack variables in the loss function is assessed and it is verified that combining these terms together results in a significant reduction of the prediction error. Moreover, the proposed model is able to capture the non-trivial dynamics of the quadrotor-slung-load system and generalizes well for a longer prediction horizon than the one considered during training. The inference times also corroborate the feasibility of the approach in a practical setting for different horizons.

Further research includes expanding this approach to a broader range of vehicles and dynamic systems. The design of model-based control strategies that leverage the proposed prediction technique is a relevant line of research to pursue, with special emphasis on long prediction horizon scenarios.







\section*{Acknowledgment}
The authors gratefully acknowledge Chrysoula Zerva and André F. T. Martins for their suggestions to improve the quality of this work.

\addtolength{\textheight}{-0cm}   




%





\end{document}